\documentclass[10pt, conference, letterpaper]{IEEEtran}
\IEEEoverridecommandlockouts

\usepackage{cite}
\usepackage{amsmath,amssymb,amsfonts}
\usepackage{algorithmic}
\usepackage{graphicx}
\usepackage{textcomp}
\usepackage{xcolor}

\usepackage{acro}
\usepackage{booktabs}
\usepackage{url}
\usepackage{siunitx}
\usepackage[hidelinks]{hyperref}

\def\BibTeX{{\rm B\kern-.05em{\sc i\kern-.025em b}\kern-.08em
    T\kern-.1667em\lower.7ex\hbox{E}\kern-.125emX}}
\begin{document}

\title{
Precision-Varying Prediction (PVP):
Robustifying ASR Systems Against Adversarial Attacks
}

\author{\IEEEauthorblockN{1\textsuperscript{st} Matías Pizarro}
\IEEEauthorblockA{\textit{Faculty of Computer Science} \\
\textit{Ruhr University Bochum}\\
Bochum, Germany \\
}
\and
\IEEEauthorblockN{2\textsuperscript{nd} Raghavan Narasimhan}
\IEEEauthorblockA{\textit{Faculty of Computer Science} \\
\textit{Ruhr University Bochum}\\
Bochum, Germany \\
}
\and
\IEEEauthorblockN{3\textsuperscript{rd} Jonas Killian}
\IEEEauthorblockA{\textit{Faculty of Computer Science} \\
\textit{Ruhr University Bochum}\\
Bochum, Germany \\
}
\and
\IEEEauthorblockN{4\textsuperscript{th} Asja Fischer}
\IEEEauthorblockA{\textit{Faculty of Computer Science} \\
\textit{Ruhr University Bochum}\\
Bochum, Germany \\
}
}

\maketitle

\thispagestyle{empty}
\pagestyle{empty}

\begin{abstract}
With the increasing deployment of automated and agentic systems, ensuring the adversarial robustness of automatic speech recognition (ASR) models has become highly relevant. 
We observe that changing the precision of an ASR model during inference reduces the likelihood of adversarial attacks to succeed.
We take advantage of this fact to make models more robust simply by randomly sampling the precision during prediction. 
Moreover, this insight can be turned into an adversarial example detection strategy by implementing a simple Gaussian classifier that thresholds the differences between outputs of models run with different precision.
To further enhance security boundaries, we combine the approach with an existing uncertainty-based defense mechanism, which forces adaptive adversaries to introduce highly perceptible noise to bypass detection.
An experimental analysis across various ASR models, languages, and attack types demonstrates a significant increase in adversarial robustness, competitive detection capabilities, and resistance to adaptive threats.
\end{abstract}


\section{Introduction}
Over the past decade, automatic speech recognition (ASR) systems have advanced rapidly, enabling voice-driven interactions that range from simple command execution in virtual assistants to open-ended conversational queries~\cite{chiu_2018_icassp}.
State-of-the-art ASR models are predominantly based on deep neural networks (DNNs)~\cite{kheddar2024Elsevier, li2025Nature}.
As these architectures are increasingly integrated into practical and safety-critical applications---including smart home systems~\cite{sharif_journal_2020}, autonomous driving~\cite{shuangshuang_2022_iccnea, caldwell2025TCSM}, and healthcare~\cite{elhadad2025Elsevier, le2025ACL, tawil_2026_icassp}---ensuring their reliability under real-world conditions is essential~\cite{williams_2022_acm}.
Given that multiple studies~\cite{du2020ACM, goodfellow2015iclr, carlini2018IEEE, qin2019PMLR, schonherr2019NDSS} demonstrate that ASR models can be maliciously manipulated via carefully designed input perturbations, establishing systemic defenses against these vulnerabilities has become a non-optional requirement for the security and trustworthiness of speech-driven technologies.

Research on securing ASR systems against adversarial threats generally splits into two directions: improving adversarial robustness or designing detectors.
The former focuses on enhancing the system's intrinsic ability to withstand adversarial inputs, typically through input transformations~\cite{pizarro_2021_ISCA, hussain_2021_USENIX} or adversarial training~\cite{madry_2018_ICLR}.
Despite their defensive intent, both strategies often degrade performance on benign inputs~\cite{wu_2023_ICLR} and suffer from distinct operational limitations.
Specifically, input transformations can introduce latency and perceptual audio artifacts, and they may lose effectiveness once they are incorporated into the attacker’s optimization process~\cite{thorsten_2021_USENIX}.
In contrast, adversarial training introduces scalability challenges due to the high computational cost of generating adversarial examples (AEs) for large-scale data~\cite{zhang_2018_ICLR}.

On the detection side, strategies deploy external detectors that operate as classifiers designed to distinguish benign from adversarial inputs.
These methods vary in their underlying mechanisms and operational constraints.
For instance, Noise Flooding~\cite{rajaratnam_2018_ISSPIT} requires multiple ASR queries to estimate the level of random noise needed to change model predictions, making it computationally expensive.
Temporal-Dependency~\cite{yang_2018_ICLR} instead analyzes sequential structure by quantifying the discrepancy---measured via Word Error Rate (WER)---between the model's prediction for a short input segment and the entire sequence, though this introduces a strict reliance on a minimum audio length.
To bypass these constraints, recent approaches look inside the network to quantify predictive uncertainty directly~\cite{daubener_2020_Interspeech, pizarro_2024_UAI}; for example, \mbox{DistriBlock}~\cite{pizarro_2024_UAI} leverages softmax distributions to compute characteristics like mean entropy for AE detection.
Despite their architectural differences, all these detection methods remain vulnerable to adaptive attacks, where an adversary with full knowledge of the defense integrates its detection logic directly into their optimization algorithm to bypass detection~\cite{zhang_2020_IJCAI}.

To address these challenges, we introduce \textit{Precision-Varying Prediction} (PVP).
Our approach leverages a novel insight: the numerical precision configurations of modern ASR models induce systematic behavioral differences that hinder the transferability of adversarial perturbations across settings.
Exploiting this phenomenon, PVP provides a dual-defense mechanism.
First, it enhances intrinsic adversarial robustness during inference by dynamically varying the model's numerical precision.
Second, it serves as a lightweight detector by comparing text transcription discrepancies across multiple precision settings to flag AEs.
Because PVP simply adjusts numerical precision settings during inference, it requires no additional training, avoids inspecting intermediate model states, and preserves benign accuracy---resulting in a highly efficient solution for Green AI deployment.
Our main contributions are the following:
\begin{itemize}
    \item \textbf{Adversarial Robustness For Free:} 
    We demonstrate that changing precision at runtime increases the adversarial robustness for free, requiring no retraining or structural changes.
    \item \textbf{Lightweight, Model-Agnostic Detection:}
    We introduce an AE detection strategy that identifies AEs simply by comparing transcription text across different precisions.
    Instead of heavy neural network training, it relies on a lightweight statistical fitting phase.   
    \item \textbf{Strategic Hybrid Integration:} 
    We show that combining PVP with existing defenses disrupts adaptive attacks.
    This combined defense forces the adversary to add much larger, audible noise into the audio to succeed, rendering the attack easily detectable by human hearing.
\end{itemize}

\section{Adversarial Attacks \& Numerical Precision}

In the following, we briefly discuss adversarial attacks on ASR systems as well as works leveraging numerical precision to increase robustness against AEs.

\subsection{Adversarial Attacks}
Adversarial attacks are methods that carefully design (ideally) imperceptible perturbations to input data that cause a machine learning model to produce incorrect outputs.

\noindent \textbf{Carlini \& Wagner Attack (C\&W)~\cite{carlini2018IEEE}:}
This iterative algorithm generates AEs by solving a constrained optimization problem via gradient descent~\cite{rumelhart_1986_nature, szegedy_2014_iclr}.
Its goal is to find an adversarial perturbation $\delta$ which, when added to the original audio $x$, misleads the ASR model $f(\cdot)$ into transcribing a malicious target phrase $y_{\text{t}}$.
Formally, the attack minimizes the following objective function:
\begin{equation*}
    \min_{\delta} \, c \cdot \mathcal{L}\big(f(x + \delta), y_{\text{t}}\big) +  \|\delta\|_2^2 \quad \text{s.t.} \quad \|\delta\|_{\infty} \le \epsilon \enspace,
    \label{eq:cw_loss}
\end{equation*}
where $\mathcal{L}$ tracks the transcription error, $\|\delta\|_2^2$ limits overall total distortion, and $\|\delta\|_{\infty} \le \epsilon$ constrains the maximum value of the perturbation.
Parameter $c$ balances success against distortion.

\noindent \textbf{Psychoacoustic Attack~\cite{qin2019PMLR, schonherr2019NDSS}:} 
This attack extends the C\&W framework by explicitly accounting for human auditory perception, producing perturbations that are largely imperceptible.
Specifically, a differentiable psychoacoustic loss penalizes perturbations that exceed masking thresholds in the time–frequency domain, ensuring that the crafted noise remains below human perceptual limits while still steering the ASR model toward a target transcription.

\subsection{Revisiting Numerical Precision for Robustness}
Some strategies leveraging numerical precision for adversarial robustness have been studied in the image domain.
For example, Sen et al.~\cite{sen_2020_empir} robustify image classifiers using an ensemble of separate, mixed-precision models ranging from full FP32 down to 4-bit configurations.
Alternatively, Fu et al.~\cite{fu_2021_PMLR} store a single model and dynamically switch to lower integer bit-widths at inference time by clipping the most significant bits via a bit-shift operation.

However, translating these image-centric paradigms to the speech domain introduces structural and acoustic limitations.
First, ensemble-based defenses are impractical for ASR systems; running multiple concurrent architectures simultaneously requires excessive GPU memory and runtime overhead.
Furthermore, combining their variable-length sequence outputs is not trivial, as it requires complex text-alignment algorithms to handle token discrepancies between different model transcriptions.
Second, dynamic integer-space truncation already degrades performance on benign inputs within the image domain.
Because ASR models process sequential inputs over time, such lossy bit-clipping would introduce quantization noise, degrading baseline accuracy. 

\section{Approach}
To address these limitations, we propose \textit{Precision-Varying Prediction} (PVP), a method that modifies precision within native hardware floating-point formats, maintaining low computational overhead while preserving performance on benign data.
The core idea is that adversarial perturbations exhibit reduced stability when the numerical precision of the model is varied at inference.
We leverage this behavior first to inherently boost the model's adversarial robustness without additional training costs.
Then, by quantifying transcription discrepancies across different precision modes, we construct a simple, lightweight attack detector.

\subsection{ASR Under Varying Numerical Precision}
Modern deep learning frameworks support multiple floating-point formats—FP32 (32-bit single precision with a larger mantissa and dynamic range) and reduced-precision formats such as FP16 (16-bit, smaller mantissa and narrower range) and BF16 (16-bit with an FP32-like exponent but a reduced mantissa). 
While FP32 provides higher numerical stability, FP16 and BF16 improve computational efficiency and memory usage.
In practice, models rarely execute in a single precision.
Instead, runtime behavior results from software-level automatic mixed precision~\cite{pytorch_amp_ops}, which applies dynamic casting and gradient scaling, and hardware/backend constraints~\cite{nvidia_cublas,nvidia_cudnn_frontend_docs}, where operations such as matrix multiplications may use reduced-precision inputs but accumulate in higher precision (often FP32)~\cite{fasi2021numerical}.
We define the exposed precision controls as the user-configurable storage dtype, autocast compute dtype, and activation of gradient scaling.
Distinguishing storage from compute precision and training from inference precision, we manipulate only the exposed compute precision as a deployment-level control variable.
This induces systematic variations in model behavior without retraining or architectural modifications, allowing us to evaluate its impact on adversarial robustness.

\subsection{Precision Sensitivity of Adversarial Examples (AEs)}
For clarity, we represent the ASR system as a single function mapping an input speech signal to a textual transcription, abstracting away intermediate components such as acoustic modeling and decoding.
Let $f_p(\cdot)$ denote an ASR model evaluated with numerical precision $p \in \mathcal{P}$, where $\mathcal{P}$ is the set of all precision configurations considered.
Let $x$ be a benign input and $\tilde{x}$ an AE crafted with a source precision $p_s$.
By construction, the AE alters the model prediction under that precision, i.e., 
$
f_{p_s}(\tilde{x}) \neq f_{p_s}(x).
$
We hypothesize that AEs are more sensitive to changes in numerical precision than benign inputs.
To analyze this, we introduce an alternative inference precision $p_a \neq p_s$.
Under this assumption, evaluating the same adversarial input across different precisions may lead to inconsistent outputs:
$
f_{p_s}(\tilde{x}) \not\approx f_{p_a}(\tilde{x}),
$
whereas benign inputs are expected to produce more stable transcriptions across precision settings:
$
f_{p_s}(x) \approx f_{p_a}(x).
$
This hypothesized differential stability across precision modes forms the basis for both a free increase in the robustness of ASR systems and a simple detection strategy.

\subsection{Robustness via Stochastic Precision Sampling}
If our hypothesis holds true, this allows for a very simple way of increasing the robustness of ASR systems, namely by simply drawing the numerical precision randomly during inference.
The induced numerical variability will make those AEs less effective that have been optimized for another precision setting.

\subsection{Detection via Precision-Diversity Scoring}
To gain a simple attack detection mechanism, we first
evaluate the ASR model under multiple precision settings $\{p_1, \dots, p_K\} \subset \mathcal{P}$, where $K$ is the number of precision configurations considered, and obtain a set of transcriptions: $\mathcal{Y}(x) = \{ f_{p_1}(x), \dots, f_{p_K}(x) \}.$
We then quantify transcription consistency using a similarity measure $s(\cdot, \cdot)$.

In practice, we instantiate $s$ using the WER, although other sequence dissimilarity metrics could be used.
To capture overall robustness to precision variation, we compute the average pairwise dissimilarity across all precision combinations and refer to it as the precision-diversity score:
\begin{equation}
D(x) = \frac{2}{K(K-1)}
\sum_{1 \le i < j \le K}
\mathrm{s}\big(f_{p_i}(x), f_{p_j}(x)\big) \enspace.
\label{eq:score}
\end{equation}
Higher precision-diversity scores indicate greater sensitivity to precision changes.
We model the distribution of precision-diversity scores for benign data using a set of 
benign samples $\mathcal{X}_{\text{b}} = \{x^{(i)}\}_{i=1}^N$.
For each $x^{(i)} \in \mathcal{X}_{\text{b}}$, we compute $D(x^{(i)})$
and fit a Gaussian distribution: $D(x) \sim \mathcal{N}(\mu, \sigma^2).$
An unknown input is then classified as adversarial if its precision-diversity score
deviates significantly from the benign distribution.

\subsection{Adaptive Attacks}
We further evaluate our strategies against defense-aware adversaries by implementing a multi-precision adaptive variant of the C\&W attack.
Rather than optimizing the objective under a single inference precision, the perturbation is jointly optimized across all precisions in $\mathcal{P}$. 
As the detector operates on WER, differences computed after decoding—an inherently non-differentiable process—the attacker cannot directly optimize the detection score. 
Consequently, the adaptive attack relies on a differentiable surrogate objective defined over precision-specific forward passes:
\begin{equation}
\min_{\delta} \quad \lVert \delta \rVert_q 
+ c \cdot \sum_{p \in \mathcal{P}} \frac{1}{|\mathcal{P}|} \cdot
\mathcal{L}\big(f_p(x + \delta), y_{\text{t}}\big) \enspace.
\label{eq:adaptive}
\end{equation}

\subsection{Hybrid Defense Framework}
We extend our approach into a hybrid framework by integrating PVP with \mbox{DistriBlock}~\cite{pizarro_2024_UAI}, a state-of-the-art defense that detects AEs using token distribution characteristics as a proxy for ASR uncertainty.
Specifically, we leverage the mean entropy, which has proven to be a highly effective characteristic for distinguishing benign from adversarial inputs.
This metric is computed by calculating the Shannon entropy~\cite{shannon_1948_journal} over the token probability distribution at each individual output time step, and subsequently taking the average.

We evaluate this hybrid framework under two distinct feature configurations: (i) $\text{Hybrid}_{\text{JS}}$ (Joint Score), which forms a two-dimensional feature representation by combining the single mean entropy score with our precision-diversity score, and (ii) $\text{Hybrid}_{\text{MP}}$ (Multi-Precision), which constructs a feature vector by concatenating the mean entropy across all precision settings.
Using these feature vectors, we estimate the empirical mean and covariance matrix from the benign reference data to fit a Multivariate Normal (MVN) distribution.
At inference time, an input is flagged as adversarial if its log probability density deviates significantly from this MVN distribution.
\begin{table*}[!t]
\caption{
Cross-precision evaluation of ASR systems on benign LibriSpeech data.
The first column indicates the training precision of each model, while the remaining columns report performance under different inference precisions.
Results are reported as WER / SER on the test-clean and test-other subsets, where lower values indicate better recognition performance.
$^{\dagger}$
}
\label{tab:asr_precision}
\centering
\scriptsize
\begin{tabular}{l|ll|ll|ll|ll}
\toprule
\multicolumn{1}{c|}{ASR Model -}  
& \multicolumn{2}{c|}{FP32} 
& \multicolumn{2}{c|}{FP16} 
& \multicolumn{2}{c|}{BF16}
& \multicolumn{2}{c}{Random sampling precision}\\
Training precision &  Test-clean & Test-other 
& Test-clean & Test-other 
& Test-clean & Test-other & Test-clean & Test-other \\
\midrule
CTC-FP32 & \underline{02.04 / 25.50} & \underline{04.02 / 40.42} & 02.04 / 25.46 & 04.03 / 40.46 & 02.05 / 25.57 & 04.03 / 40.46 & 02.04 / 25.51 & 04.02 / 40.44 \\
CTC-FP16 & 01.98 / 25.15 & 04.00 / 40.18 & \underline{01.98 / 25.19} & \underline{03.99 / 40.08} & 02.00 / 25.15 & 04.01 / 40.22 & 01.98 / 25.16 & 04.00 / 40.16 \\
CTC-BF16 & 01.91 / 24.69 & 03.93 / 40.08 & 01.91 / 24.73 & 03.93 / 40.08 & \underline{01.92 / 24.73} & \underline{03.95 / 40.15} & 01.91 / 24.71 & 03.93 / 40.10 \\
\midrule
seq2seq-FP32 & \underline{02.80 / 31.49} & \underline{08.33 / 56.75} & 02.80 / 31.56 & 08.30 / 56.99 & 03.49 / 38.44 & 09.28 / 61.89 & 03.03/ 33.83 & 08.63 / 58.54 \\
seq2seq-FP16 & 02.87 / 31.49 & 08.64 / 57.98 & \underline{02.92 / 31.95} & \underline{08.72 / 57.98} & 05.75 / 59.39 & 11.97 / 74.04 & 03.84 / 40.94 & 09.77 / 63.33 \\
seq2seq-BF16 & 02.65 / 30.00 & 07.99 / 54.37 & 02.66 / 29.96 & 07.97 / 54.41 & \underline{02.68 / 29.81} & \underline{07.94 / 54.37} & 02.66 / 29.92 & 07.96 / 54.38 \\
\midrule
Transformer-FP32 & \underline{02.15 / 26.11} & \underline{05.12 / 43.35} & 02.15 / 26.07 & 05.12 / 43.38 & 02.17 / 26.30 & 05.11 / 43.52 & 02.15 / 26.16 & 05.11 / 43.41 \\
Transformer-FP16 & 02.15 / 25.69 & 05.17 / 44.57 & \underline{02.15 / 25.69} & \underline{05.16 / 44.54} & 02.16 / 25.80 & 05.14 / 44.37 & 02.15 / 25.72 & 05.15 / 44.49 \\
Transformer-BF16 & 02.18 / 26.34 & 05.06 / 44.27 & 02.18 / 26.37 & 05.05 / 44.30  & \underline{02.18 / 26.37} & \underline{05.05 / 44.13} & 02.18 / 26.36 & 05.05 / 44.23 \\
\midrule
Whisper-FP32 & \underline{02.03 / 24.08} & \underline{04.74 / 41.61} & 02.03 / 24.05 & 04.74 / 41.61 & 02.03 / 24.12 & 04.73 / 41.51 & 02.03 / 24.08 & 04.73 / 41.57 \\
Whisper-FP16 & 02.05 / 24.54 & 04.77 / 41.65 & \underline{02.05 / 24.47} & \underline{04.76 / 41.61} & 02.05 / 24.47 & 04.75 / 41.48 & 02.05 / 24.49 & 04.76 / 41.58 \\
Whisper-BF16 & 02.03 / 24.12 & 04.70 / 41.31 & 02.03 / 24.16 & 04.70 / 41.37 & \underline{02.05 / 24.39} & \underline{04.71 / 41.44} & 02.03 / 24.22 & 04.70 / 41.37 \\ 
\bottomrule
\multicolumn{8}{l}{$^{\dagger}$Underline denotes matched training and inference precision.}
\end{tabular}
\end{table*}
\section{Experiments and Results}
All experiments are conducted under three precision configurations: FP32, FP16, and BF16.
Our adversarial attack implementation follows~\cite{olivier_2022_interspeech}, and all models and hyperparameters are available in our repository: \url{https://github.com/blindconf/multi_precision_fusion}.

\subsection{ASR Models}
We evaluate our approach across four distinct ASR architectures: CTC, seq2seq, Transformer, and Whisper, using the official SpeechBrain recipes~\cite{ravanelli2021speechbrain}.
To test the language generalization of our proposal, we use both English and German speech data.
Specifically, all four architectures are trained on the English LibriSpeech dataset~\cite{panayotov2015IEEE}, while the CTC and seq2seq models are also trained on the Mozilla Common Voice German Scripted Speech 25.0~\cite{ardila-etal-2020-common}.
Each available model is trained separately under three distinct precision modes, resulting in a total of 18 trained models.
The architectural configurations are structured as follows:

\noindent \textbf{CTC:} A model built upon a pretrained wav2vec~2.0-based encoder~\cite{Baevski_2020_NEURIPS} trained using the Connectionist Temporal Classification (CTC) loss~\cite{Graves_2006_ICML}.

\noindent \textbf{seq2seq:} An encoder–decoder architecture combining convolutional and recurrent layers with attention-based decoding~\cite{William_2016_ICASSP} trained using a joint CTC and negative log-likelihood (NLL) loss.

\noindent \textbf{Transformer:} A fully attention-based encoder–decoder architecture trained with a joint CTC and NLL loss, which uses a pretrained Transformer language model from SpeechBrain~\cite{Wolf_2020_etal} during decoding.

\noindent \textbf{Whisper:} A pretrained sequence-to-sequence ASR system developed by OpenAI~\cite{radford2023PMLR}, optimized using an NLL loss.

\subsection{Evaluation Metrics}
To evaluate ASR performance, we report two standard metrics: the WER and Sentence Error Rate (SER)~\cite{pizzi_2024_spsc}. 
The WER is an alignment-based metric that compares a hypothesis to a reference transcription using the Levenshtein distance~\cite{Navarro_2001_ACM}.
Errors are defined as insertions, deletions, and substitutions required to transform the hypothesis into the reference computed at the word level.
The SER evaluates errors at the utterance level: a sentence is counted as incorrect if the hypothesis differs from the reference by at least one such transcription error.
To quantify distortion in AEs, we use the Segmental Signal-to-Noise Ratio~($\text{SNR}_{seg}$) that is computed by averaging the frame-wise energy ratios and aligns more closely with human auditory perception than the non-segmental measure~\cite{Mermelstein_1979_ASA}.

\subsection{Precision Variation on Benign Data Across ASR Systems}
\label{sec:variation_precision}
\noindent \textbf{Experimental Setup:}
All models were trained using a fixed numerical precision and were evaluated under two settings: (i) cross-precision inference, where models were tested using alternative fixed precisions, and (ii) stochastic precision, where the precision was randomly sampled at prediction time (with results averaged over $10$ trials).
Experiments were conducted across two languages.
For English, we used benign speech data from the \textit{test-clean} (high-quality recordings) and \textit{test-other} (acoustically challenging recordings) datasets from LibriSpeech.
For German, we used the benign speech samples from the test split of the Mozilla Common Voice corpus.
Results were quantified using the WER and SER, where the hypothesis corresponds to the ASR model output and the reference is the ground-truth transcription.

\noindent \textbf{Results:}
When evaluated under both cross-precision and stochastic precision settings on benign inputs, the WER and SER deviations remain minimal compared to standard fixed-precision baselines.
This trend holds across architectures and languages: differences are negligible on LibriSpeech (Tab.~\ref{tab:asr_precision}) as well as on the Mozilla Common Voice corpus (Tab.~\ref{tab:asr_precision_ge}).
Across all models, these results demonstrate that changes in numerical format alone do not degrade baseline ASR performance, proving that precision-varying inference maintains standard accuracy without language-specific degradation.
\begin{table}[!t]
\caption{
Cross-precision evaluation on the Mozilla Common Voice German test set. 
Columns show inference configurations across different precisions.
Results given as WER / SER, where lower values indicate better recognition performance.~$^{\dagger}$
}
\label{tab:asr_precision_ge}
\centering
\resizebox{\columnwidth}{!}{
\begin{tabular}{l|l|l|l|l}
\toprule
ASR Model & FP32 & FP16 & BF16 & Random sampling \\
\midrule
CTC-FP32
& {\underline{09.19 / 43.50}} & {09.18 / 43.47} & {09.18 / 43.42} & 09.18 / 43.46 \\
CTC-FP16 & {08.93 / 42.74} & {\underline{08.93 / 42.71}} & {08.95 / 42.79} & {08.93 / 42.74} \\
CTC-BF16 & {09.13 / 43.48} & {09.13 / 43.48} & {\underline{09.12 / 43.40}} & {09.13 / 43.45} \\
\midrule
seq2seq-FP32 & {\underline{12.95 / 52.03}} & {12.94 / 51.96} & {13.15 / 52.43} & {13.01 / 52.14} \\
seq2seq-FP16 & {13.01 / 52.85} & {\underline{13.21 / 52.87}} & {13.86 / 53.75} & 13.36 / 53.15 \\
seq2seq-BF16 & {13.11 / 52.24} & {13.10 / 52.20} & {\underline{12.93 / 52.81}} & {13.04 / 52.41} \\
\bottomrule
\multicolumn{5}{l}{$^{\dagger}$Underline denotes matched training and inference precision.}
\end{tabular}
}
\end{table}
\begin{table*}[!htbp]
\caption{
Adversarial robustness evaluation of ASR systems under PVP. 
Columns show inference configurations across different precisions.
For each attack, the final column reports distortion ($\text{SNR}_{\text{seg}}$ in \si{\decibel}); all other entries show WER / SER.
Results are evaluated on 100 AEs generated per attack type from the LibriSpeech test-clean set.
}
\label{tab:attack_precision}
\centering
\resizebox{\textwidth}{!}{
\begin{tabular}{l|lllll|lllll}
\toprule
 \multicolumn{1}{c|}{ASR Model - } & \multicolumn{5}{c|}{C\&W attack} & \multicolumn{5}{c}{Psychoacoustic attack} \\
Training precision 
 & FP32 & FP16 & BF16 & Random {\color{gray}$\uparrow$} & $\text{SNR}_{seg}$ 
 & FP32 & FP16 & BF16 & Random {\color{gray}$\uparrow$} & $\text{SNR}_{seg}$ \\
\midrule
CTC-FP32 
& \underline{00.00 / 00.00} & 06.38 / 28.00 & 25.72 / 83.00 & 10.93 / 38.90 & 18.54
& \underline{00.00 / 00.00} & 08.02 / 36.00 & 25.31 / 84.00 & 10.74 / 39.30 & 19.94 \\
CTC-FP16 
& 03.29 / 15.00 & \underline{00.00 / 00.00} & 25.93 / 83.00 & 10.29 / 34.50 & 17.94
& 09.47 / 39.00 & \underline{00.00 / 00.00} & 30.25 / 89.00 & 13.81 / 44.00 & 19.37\\
CTC-BF16 
& 19.55 / 70.00 & 18.93 / 67.00 & \underline{00.00 / 00.00} & \textbf{12.59 / 44.80} & 17.53
& 20.78 / 81.00 & 21.40 / 83.00 & \underline{00.00 / 00.00} & \textbf{13.87 / 53.30} & 18.79\\
\midrule
seq2seq-FP32 
& \underline{00.00 / 00.00} & 22.02 / 48.00 & 34.16 / 77.00 & 18.50 / 41.20 & 13.34 
& \underline{00.00 / 00.00} & 27.16 / 58.00 & 35.19 / 79.00 & 21.52 / 47.80 & 13.85 \\
seq2seq-FP16 
& 56.38 / 95.00 & \underline{00.41 / 01.00} & 62.35 / 97.00 & \textbf{40.58 / 66.20} & 15.89 
& 57.41 / 98.00 & \underline{00.21 / 01.00} & 60.70 / 93.00 & \textbf{40.21 / 65.90} & 16.33 \\
seq2seq-BF16 
& 50.41 / 98.00 & 51.85 / 98.00 & \underline{00.00 / 00.00} & 33.79 / 64.10 & 17.23 
& 50.62 / 97.00 & 50.41 / 97.00 & \underline{00.00 / 00.00} & 33.15 / 63.30 & 17.81 \\
\midrule
Transformer-FP32 
& \underline{00.00 / 00.00} & 24.90 / 35.00 & 44.44 / 57.00 & 23.85 / 31.30 & 28.42
& \underline{00.00 / 00.00} & 06.58 / 07.00 & 24.28 / 28.80 & \textbf{11.15} / 13.00 & 24.84 \\
Transformer-FP16 
& 22.63 / 40.00 & \underline{00.00 / 00.00} & 35.19 / 56.00 & 19.34 / 32.80 & 28.42
& 07.41 / 09.00 & \underline{00.00 / 00.00} & 15.23 / 26.00 & 07.49 / 12.40 & 24.23 \\
Transformer-BF16 
& 58.44 / 86.00 & 19.75 / 30.00 & \underline{00.00 / 00.00} & \textbf{37.45 / 55.20} & 28.27
& 16.46 / 25.00 & 16.46 / 25.00 & \underline{00.00 / 00.00} & 10.31 / \textbf{14.90} & 25.04 \\
\midrule
Whisper-FP32 
& \underline{00.00 / 00.00} & 79.48 / 72.00 & 66.14 / 61.00 & 47.25 / 43.80 & 26.37
& \underline{00.00 / 00.00} & 60.76 / 52.00 & 35.06 / 29.00 & 26.89 / 23.00 & 23.88 \\
Whisper-FP16 
& 28.69 / 31.00 & \underline{10.36 / 08.00} & 44.42 / 43.00 & 27.91 / 27.50 & 31.86
& 58.37 / 60.00 & \underline{10.36 / 08.00} & 59.76 / 65.00 & \textbf{44.16} / \textbf{45.80} & 33.23 \\
Whisper-BF16 
& 63.35 / 63.00 & 92.63 / 88.00 & \underline{00.00 / 00.00} & \textbf{52.09 / 50.60} & 26.30
& 22.31 / 21.00 & 59.36 / 53.00 & \underline{00.00 / 00.00}  & 26.97 / 24.90 & 23.72\\
\bottomrule
\multicolumn{11}{l}{$^{\dagger}$Underline denotes matched training and inference precision; bold marks the highest WER / SER under random precision sampling (strongest adversarial robustness).}
\end{tabular}
}
\end{table*}
\subsection{Impact of Precision Variation on Adversarial Robustness}
\label{sec:Robustness_precision}
\noindent \textbf{Experimental Setup:}
We randomly selected $100$ benign samples from the LibriSpeech test-clean dataset and $100$ samples from the Mozilla Common Voice test set.
For LibriSpeech, we generated corresponding AEs using both the C\&W and psychoacoustic attacks, whereas for Common Voice, we generated AEs using the C\&W attack.
In all cases, we ran $4{,}000$ optimization iterations per sample.
For each sample, a distinct adversarial target transcript was randomly drawn from the respective dataset, ensuring that no two samples shared the same target transcription.
Psychoacoustic AEs were initialized from the C\&W-generated examples, beginning from inputs that already fooled the respective ASR model.
Attack success was quantified using the WER and SER between the ASR output and the target transcript, where a score of zero denotes perfect success.
Perceptual distortion was measured via the $\text{SNR}_{seg}$, where higher values correspond to less perceptible noise.
To evaluate adversarial robustness, we tested whether the adversarial effect persists under changes in inference precision, tracking attack degradation across the cross-precision and stochastic-precision settings.

\noindent \textbf{Results:}
When attack generation and evaluation were performed with matching precision, the WER and SER remained near zero, confirming highly successful attacks across both English (Tab.~\ref{tab:attack_precision}) and German (Tab.~\ref{tab:attack_precision_ge}) benchmarks.
The measured $\text{SNR}_{seg}$ values remain consistently high for all ASR models, indicating low perceptual distortion of the adversarial examples (values consistent with those reported by \cite{pizzi_2026_CSL}).
However, changing the inference precision (either deterministically or randomly) consistently increased both the WER and SER across all evaluated architectures.
Notably, models trained under the BF16 format tend to exhibit the highest adversarial robustness under a precision mismatch. 
Because BF16 preserves the wide dynamic range of FP32 with significantly reduced fraction precision, adversarial perturbations optimized to exploit this numerical spacing become unstable when forced to infer within the finer precision grids of FP16 or FP32.
These empirical outcomes hold across both benchmarks, reinforcing our conclusion that precision-induced instability provides a reliable, language-agnostic layer of adversarial robustness.
\subsection{Adversarial Detection and Hybrid Defense Evaluation}
\label{sec:detecting_aes}
\noindent \textbf{Experimental Setup:}
To construct the benign reference distributions, we randomly sampled $200$ LibriSpeech utterances ($100$ from test-clean and $100$ from test-other) and $100$ utterances from the Mozilla Common Voice test set.
For the English benchmark, we evaluated standalone defenses, comparing our PVP approach against Noise Flooding (NF)~\cite{rajaratnam_2018_ISSPIT}, Temporal Dependency (TD)~\cite{yang_2018_ICLR}, and \mbox{DistriBlock}~\cite{pizarro_2024_UAI}.
For the German benchmark, we evaluated standalone PVP against \mbox{DistriBlock} alongside our proposed hybrid defense frameworks ($\text{Hybrid}_{\text{JS}}$ and $\text{Hybrid}_{\text{MP}}$).

For a fair comparison among the standalone defenses, a separate Gaussian distribution was fitted for each ASR model.
These distributions were based on the individual defense scores of PVP, NF, TD, and DistriBlock obtained from the respective benign reference data.
For the multi-featured hybrid defenses on the German benchmark, we fitted an MVN distribution whose empirical mean vector and covariance matrix were estimated using the combined feature vectors computed from the same $100$ German benign reference samples.
Detection performance was measured using the Area Under the Receiver Operating Characteristic curve (AUROC).
The English evaluation was conducted on $200$ AEs and a disjoint set of $200$ benign samples, while the German evaluation set comprised $100$ AEs and a disjoint set of $100$ benign samples.

\begin{table}[!t]
\caption{
Adversarial robustness evaluation of ASR systems under PVP. 
Columns show inference configurations across different precisions.
The final column reports distortion, while other entries show WER / SER.
Results are evaluated on 100 C\&W AEs generated from the Mozilla Common Voice test set.
Bold marks the highest WER / SER under random sampling.~$^{\dagger}$
}
\label{tab:attack_precision_ge}
\centering
\resizebox{\columnwidth}{!}{
\begin{tabular}{l|l|l|l|l|l}
\toprule
ASR Model & FP32 & FP16 & BF16 & Random {\color{gray}$\uparrow$} & $\text{SNR}_{seg}$ \\
\midrule
CTC-FP32   & \underline {00.00 / 00.00}  & 09.74 / 41.00  & 28.57 / 89.00 & \textbf{12.77} / 43.33 & 18.96 \\
CTC-FP16   & 08.66 / 35.00 & \underline {00.00 / 00.00}   & 25.76 / 87.00 & 11.47 / 40.66 & 19.74 \\
CTC-BF16   & 18.61 / 75.00  & 19.26 / 77.00 & \underline {00.00 / 00.00} & 12.62 / \textbf{50.67} & 20.04 \\
\midrule
seq2seq-FP32 & \underline{02.81 / 03.00}  & 23.81 / 44.00 & 39.83 / 75.00 & 22.15 / 40.66 & 10.96 \\
seq2seq-FP16 & 44.37 / 80.00 & \underline{00.65 / 01.00}   & 48.27 / 82.00 & 31.09 / 54.30 & 11.31 \\
seq2seq-BF16 & 50.65 / 91.00 & 50.22 / 91.00 & \underline{01.73 / 02.00}  & \textbf {34.20 / 61.33}  & 10.51 \\
\bottomrule
\multicolumn{6}{l}{$^{\dagger}$Underline denotes matched training and inference precision.
}
\end{tabular}
}
\end{table}
\noindent \textbf{Results on LibriSpeech:}
Tab.~\ref{tab:ae_detection} demonstrates that precision diversity enables effective adversarial detection.
Across models and precisions, PVP achieved strong separability, yielding AUROC values above $0.90$ for CTC and seq2seq architectures, with reduced effectiveness for Transformer and Whisper architectures likely due to their large-scale pretrained designs and inherent mixed-precision optimizations.
Take, for instance, the estimated mean and standard deviation ($\mu \pm \sigma$) for the underlying PVP scores are $0.17 \pm 1.75$ for CTC-FP32, $0.11 \pm 1.57$ for Whisper-FP16, and $0.61 \pm 4.23$ for seq2seq-BF16.
This overall tight distribution near zero confirms that benign inputs generate negligible precision-diversity scores, explaining why the framework rarely misclassifies benign speech samples.

Compared to alternative defenses, PVP offers distinct practical and architectural advantages.
While NF relies on an iterative, open-ended search algorithm, PVP bounds its inference overhead to a small, fixed number of precision types, reducing the average execution time from $7.36$ to $0.08$ seconds per sample (benchmarked using a CTC model on an NVIDIA A40 GPU with $48$ GB VRAM).
Furthermore, TD completely failed on short utterances containing only one to two words (which had to be excluded from evaluation), whereas \mbox{DistriBlock} achieved high overall detection performance but strictly depended on access to full token-level distributions, limiting its applicability in restricted-access settings.

\noindent \textbf{Results on Common Voice:}
As shown in Tab.~\ref{tab:ae_detection_ge}, PVP achieved detection scores that remain consistent with the results obtained on the LibriSpeech benchmark.
However, by combining the unique characteristics of PVP with those of \mbox{DistriBlock}, the unified hybrid configurations unlocked further performance gains.
Specifically, $\text{Hybrid}_{\text{JS}}$ achieved the highest overall detection performance across all precisions, yielding near-perfect separation with AUROC scores of at least $0.98$.
While $\text{Hybrid}_{\text{MP}}$ also provided a noticeable boost over standalone PVP, it was consistently outperformed by the $\text{Hybrid}_{\text{JS}}$ formulation.
This performance enhancement highlights the complementary value of integrating precision diversity with other feature types rather than relying solely on a single detection modality.
\begin{table}[!t]
\caption{
Adversarial detection performance (AUROC) across different defense methods.
Benign samples from the LibriSpeech test-clean set are evaluated against their AE counterparts (C\&W and Psy). 
Bold marks the highest detection score.~$^{\dagger}$
}
\label{tab:ae_detection}
\centering
\resizebox{\columnwidth}{!}{
\begin{tabular}{l|l|l|llll}
\toprule
\multicolumn{1}{c|}{ASR Model -}  
& \multicolumn{1}{c|}{C\&W vs.} 
& \multicolumn{1}{c|}{Psy. vs.} 
& \multicolumn{4}{c}{Both (C\&W and Psy.) vs. benign}\\
Training precision &  benign & benign 
& PVP & NF & TD & DistriBlock \\
\midrule
CTC-FP32 & 0.92 & 0.93 & 0.92 & 0.89 & 0.94 & \textbf{0.99} \\
CTC-FP16 & 0.91 & 0.95 & 0.95 & 0.89 & 0.93 & \textbf{0.99} \\
CTC-BF16 & 0.86 & 0.91 & 0.91 & 0.87 & 0.94 & \textbf{0.99} \\
\midrule
seq2seq-FP32 & 0.90 & 0.93 & 0.91 & 0.70 & 0.83 & \textbf{0.96} \\
seq2seq-FP16 & 0.97 & 0.98 & \textbf{0.97} & 0.73 & 0.81 & \textbf{0.97} \\
seq2seq-BF16 & 0.99 & 0.98 & \textbf{0.98} & 0.71 & 0.83 & 0.96 \\
\midrule
Transformer-FP32 & 0.86 & 0.64 & 0.75 & 0.89 & 0.89 & \textbf{0.97} \\
Transformer-FP16 & 0.85 & 0.64 & 0.75 & 0.91 & 0.91 & \textbf{0.97} \\
Transformer-BF16 & 0.93 & 0.63 & 0.78 & 0.89 & 0.85 & \textbf{0.97} \\
\midrule
Whisper-FP32 & 0.93 & 0.79 & 0.86 & 0.90 & \textbf{0.95} & 0.94 \\
Whisper-FP16 & 0.70 & 0.84 & 0.77 & 0.85 & 0.91 & \textbf{0.99} \\
Whisper-BF16 & 0.95 & 0.77 & 0.86 & 0.91 & \textbf{0.96} & 0.95 \\
\bottomrule
\multicolumn{7}{l}{$^{\dagger}$C\&W and Psy use 100 benign and 100 AEs each, while Both uses 200 of each.}
\end{tabular}
}
\end{table}
\subsection{Defensive Security Boundaries Under Adaptive Attacks}
To evaluate the worst-case security boundaries of our PVP, we generated adaptive AEs optimized over $1{,}000$ iterations per sample to actively minimize the targeted defense objectives (Eq.~\ref{eq:adaptive}).
Under this threat, all attacks achieve an SER of zero, indicating complete attack success and demonstrating that the precision-diversity method can be circumvented when the adversary explicitly optimizes against it---a vulnerability shared equally by all existing defense frameworks.

However, this evasion comes at a cost to acoustic imperceptibility when facing hybrid defenses.
As reported in Tab.~\ref{tab:hybrid_detection_ge} for the Common Voice benchmark, while standalone PVP and \mbox{DistriBlock} were bypassed with relatively low distortion ($\text{SNR}_{\text{seg}}$ between $14.50$\,\si{\decibel} and $18.79$\,\si{\decibel}), bypassing the hybrid frameworks drastically degraded signal quality.
Most notably, the $\text{Hybrid}_{\text{MP}}$ forced the highest distortion levels across the board, pushing the $\text{SNR}_{\text{seg}}$ down to an audible $-0.54$\,\si{\decibel} on the FP32 model.
These empirical outcomes confirm that while a multi-objective adaptive attack can technically bypass the statistical thresholds, the hybrid framework effectively breaks the attack's primary constraint by forcing a trade-off between detection evasion and acoustic imperceptibility.
\begin{table}[!t]
\caption{
Adversarial detection performance (AUROC) across different defense methods.
$100$ Benign samples from the Common Voice German test set are evaluated against their C\&W AE counterparts. 
Bold marks the highest detection score.
}
\label{tab:ae_detection_ge}
\centering
\scriptsize
\begin{tabular}{l|l|c|l|l}
\toprule
ASR Model &  PVP & DistriBlock & $\text{Hybrid}_{\text{JS}}$ & $\text{Hybrid}_{\text{MP}}$ \\
\midrule
CTC-FP32 & 0.92 & 0.95 & \textbf{0.99} & 0.95 \\
CTC-FP16 & 0.94 & 0.96 & \textbf{0.98} & 0.95\\
CTC-BF16 & 0.86 & 0.96 & \textbf{0.98} & 0.96\\
\bottomrule
\end{tabular}
\end{table}
\begin{table}[!t]
\caption{
Adaptive attack detection distortion ($\text{SNR}_{\text{seg}}$ in \si{\decibel}) and performance (AUROC) across defense methods.
Evaluated using $100$ benign samples from the Common Voice German test set against their adaptive AE counterparts.~$^{\dagger}$
}
\label{tab:hybrid_detection_ge}
\centering
\resizebox{\columnwidth}{!}{
\begin{tabular}{c|c|c|c|c}
\toprule
ASR 
& PVP 
& DistriBlock 
& $\text{Hybrid}_{\text{JS}}$ 
& $\text{Hybrid}_{\text{MP}}$ \\
Model & $\text{SNR}_{\text{seg}}$/AUROC
& $\text{SNR}_{\text{seg}}$/AUROC
& $\text{SNR}_{\text{seg}}$/AUROC
& $\text{SNR}_{\text{seg}}$/AUROC \\
\midrule
CTC-FP32 & 18.00 / 0.48 & 14.50 / 0.69 & 08.79 / 0.63 & \textbf{-00.54} / 0.47 \\
CTC-FP16 & 17.39 / 0.48 & 15.34 / 0.68 & 10.09 / 0.63 & \textbf{03.57} / 0.50 \\
CTC-BF16 & 18.79 / 0.47 & 15.01 / 0.70 & 08.37 / 0.64 & \textbf{01.40} / 0.53 \\
\bottomrule
\multicolumn{5}{l}{$^{\dagger}$Bold marks the highest distortion among the generated AEs.}
\end{tabular}
}
\end{table}
\section{Conclusion}
We demonstrate that numerical precision serves as an effective, lightweight mechanism for improving ASR adversarial robustness. 
Across diverse architectures, languages, and attacks, adversarial inputs exhibit distinct instability when inference precision varies, while benign inputs remain consistent. 
Because varying execution precision is native to modern hardware, this protection is obtained virtually ``for free'' without additional training, architectural changes, or data augmentation.

We leverage this phenomenon via a dual-stage paradigm: first, as a stochastic sampling strategy to boost baseline robustness, and second, as a standalone detector. 
Furthermore, integrating this approach into a multi-feature hybrid defense significantly raises the security threshold.
Even though an adaptive adversary can bypass the joint constraints of these hybrid defenses, doing so forces the optimization to add perceptible noise, making the attack easy to detect.
Future work will explore how other existing defense modalities can be integrated into this hybrid framework to establish an even more comprehensive barrier against adaptive threats.
\newpage
\section*{AI-Generated Content Disclosure}
Generative AI tools were used solely for language editing, grammar correction, and improving the clarity of the manuscript.
No generative AI tools were used to produce or design the scientific content, experiments, or results presented in this paper.
All ideas, methods, and evaluations were developed by the authors, who take full responsibility for the content.
\bibliographystyle{IEEEtran}
\bibliography{paper}

\end{document}